\documentclass[11pt]{article}

\usepackage[final]{acl}

\usepackage{times}
\usepackage{latexsym}

\usepackage[T1]{fontenc}

\usepackage[utf8]{inputenc}

\usepackage{microtype}

\usepackage{inconsolata}

\usepackage{graphicx}

\usepackage{booktabs}

\usepackage[table]{xcolor}

\usepackage{subcaption}

\usepackage{makecell}

\title{A Cognitively Motivated Multidimensional Framework\\ for Evaluating Metaphor Explanations}

\author{Ana Naveriani \\
  Constructor University \\\And
  Jakob Suchan \\
  Constructor University \\\And
 Stefano Zoia \\
University of Turin \\\AND
 Mehul Bhatt \\
 Örebro University \\\And
 Antonio Lieto \\
 University of Salerno\\\And
 Gian Luca Pozzato\\
University of Turin\AND
{\bfseries CoDesign Lab EU -- Artificial and Human Intelligence}\\\href{https://codesign-lab.org}{codesign-lab.org}
}

\usepackage{xcolor}

\newcounter{mytodo}

\begin{document}
\maketitle

\begin{abstract}
Current evaluation of metaphor explanations relies mainly on holistic quality ratings, revealing little about how explanation quality is structured or where human judgments agree and diverge. We introduce a cognitively motivated framework that decomposes metaphor explanation quality into six theoretically grounded dimensions. In a dense annotation study (11,200 ratings), we find that: {\bfseries(i)} explanation quality is genuinely multidimensional; {\bfseries(ii)} annotator disagreement is systematic rather than random; and {\bfseries(iii)} the six dimensions collapse into a shared cluster and two independent axes of judgment. An exploratory feasibility study further shows that a standard automatic evaluation pipeline can recover parts of this structure, predicting the most discriminative dimensions well while its errors correlate  human (dis)agreement. Together, these results suggest that multidimensional evaluation offers richer diagnostic insight than holistic ratings, and that automatic evaluators for open-ended generation tasks should be judged on how well they preserve the structure of human judgment.
\end{abstract}

\begin{table*}[t]
\centering
\scriptsize
\definecolor{nordicbg}{RGB}{229,233,240}
\begin{tabular}{p{0.18\linewidth}p{0.26\linewidth}p{0.22\linewidth}p{0.26\linewidth}}
\rowcolor{nordicbg}
\toprule
\textbf{CRITERION} & \textbf{DEFINITION} & \textbf{THEORETICAL BASIS} & \textbf{EXAMPLE}\\
\midrule
\cellcolor{nordicbg}\textbf{NON-CIRCULARITY} & The explanation avoids restating the metaphor in equivalent figurative terms. & Principle of Experiential Grounding\newline\citep{lakoff1980metaphors} & \textcolor{gray!70!black}{Explaining \textit{Time is Money} as ``time is valuable because it is like money'' is circular.}\\
\cellcolor{nordicbg}\textbf{ORIGIN AND\newline CULTURAL ACCURACY} & The explanation captures relevant cultural, historical, or contextual background. & Cognitive Grammar \newline\citep{langacker2008cognitive} & \textcolor{gray!70!black}{\textit{Achilles' Heel} requires knowledge of its origin in Greek mythology.}\\
\cellcolor{nordicbg}\textbf{MAPPING} & The explanation identifies the conceptual relation between source and target domains. & Conceptual Metaphor Theory\newline\citep{lakoff1980metaphors} & \textcolor{gray!70!black}{\textit{Life is a Journey} maps paths, obstacles, and destinations onto stages, challenges, and goals in life.}\\
\cellcolor{nordicbg}\textbf{EMERGENCE} & The explanation captures the non-literal meaning that arises from the metaphor. & Blended Space Theory\newline\citep{fauconnier2002way} & \textcolor{gray!70!black}{\textit{Heart of Stone} conveys emotional coldness and lack of empathy, meanings that go beyond the literal properties of stone.}\\
\cellcolor{nordicbg}\textbf{ACCESSIBILITY} & The explanation is understandable to the intended audience. & Grice's Maxim of Manner\newline\citep{grice1975logic} & \textcolor{gray!70!black}{Explaining \textit{Life is a Journey} using familiar experiences is more accessible than relying on technical linguistic terminology.}\\
\cellcolor{nordicbg}\textbf{PURPOSE} & The explanation clarifies why the metaphor is communicatively effective or useful. & Grice's Maxim of Relevance\newline\citep{grice1975logic} & \textcolor{gray!70!black}{\textit{The Mind is a Computer} frames cognition as information processing, making complex mental processes easier to understand.}\\
\bottomrule
\end{tabular}
\caption{Proposed evaluation dimensions for metaphor explanations and their theoretical grounding.}
\label{tab:criteria}
\end{table*}

\section{Introduction}

Metaphors are a pervasive feature of human language and thought \citep{lakoff1980metaphors,lakoff1999philosophy}, enabling people to understand abstract concepts through more familiar and concrete experiences. 
For instance, consider the following passage:

\medskip

{\it
``Life is a journey where time is money and the mind works like a computer. But an Achilles' heel may prove difficult to overcome, and a heart of stone can keep us from what truly matters.''
}

\medskip

While most speakers intuitively understand these expressions, explaining \textit{why} they convey meaning is considerably more difficult \cite{glucksberg2001understanding}. Effective explanations must reveal underlying conceptual mappings, account for emergent meaning, connect figurative language to familiar experiences, and provide sufficient context for interpretation. Consequently, metaphor explanation is not merely a linguistic task, but also a cognitive and communicative one.
Research in cognitive linguistics has studied metaphors as a cognitive mechanism through which abstract concepts are understood via mappings from more concrete domains \citep{lakoff1980metaphors,lakoff1999philosophy}. Computational research has focused primarily on metaphor identification, interpretation, and generation \citep{shutova2010metaphor,tsvetkov2014metaphor}, while recent large language models have substantially improved the quality of ex-post generated metaphor explanations \cite{Ichien2024}. However, comparatively little work has focused on their systematic evaluation. 
As such, evaluation remains a central challenge in Natural Language Generation (NLG). Traditional automatic metrics such as BLEU \citep{papineni2002bleu}, ROUGE \citep{lin2004rouge}, and BERTScore \citep{zhang2020bertscore} often correlate poorly with human judgments for open-ended generation tasks \cite{schmidtova2024-automatic-metrics}. Consequently, recent work has increasingly emphasized human-centered and multidimensional evaluation frameworks \citep{howcroft2020twenty,celikyilmaz2021evaluation,VANDERLEE2021}, while LLM-based evaluators have emerged as promising alternatives for automatic assessment \citep{zheng2023judging,liu2023gpteval,huang2025llmjudge}.
Furthermore, human disagreement is increasingly recognized as informative factor for evaluation \cite{uma2021learning}.

Existing work on metaphor generation and explanation typically relies on holistic quality ratings or a small number of task-specific criteria \citep{chakrabarty2021mermaid,stowe2021metaphor}. Such evaluations provide limited insight into which aspects of an explanation influence human judgments or where evaluators systematically disagree. 
We address this gap by investigating how humans apply different dimensions of metaphor explanation.
Our contributions are: {\small\bfseries(1)} introducing a cognitively motivated multidimensional framework for evaluating metaphor explanations; {\small\bfseries(2)} analysing these dimensions in terms of reliability, redundancy, and disagreement; and {\small\bfseries(3)} conducting an exploratory automatic evaluation study investigating how standard language encoders can recover these aspects.

\section{A Multidimensional Framework for Evaluating Metaphor Explanations}
\label{sec:framework}

Building on insights from Conceptual Metaphor Theory \citep{lakoff1980metaphors,lakoff1999philosophy}, analogical reasoning \citep{gentner1983structure}, psycholinguistic models of metaphor comprehension \citep{glucksberg2001understanding,bowdle2005career}, and recent cognitively inspired computational approaches \citep{lieto2024delta,cappa2024onions}, we propose a multidimensional framework for evaluating metaphor explanations. The framework comprises six dimensions, capturing complementary aspects considered when assessing whether an explanation is useful, meaningful, and informative (Table \ref{tab:criteria}): \emph{Non-Circularity} evaluates informational value, \emph{Origin \& Cultural Accuracy} contextual grounding, \emph{Mapping} conceptual correspondence, \emph{Emergence} derived figurative meaning, `Accessibility' communicative clarity, and `Purpose' communicative function. In essence, the framework combines complementary constructs from cognitive linguistics and pragmatics into an operational annotation scheme rather than directly operationalizing any single theoretical account.

\section{Analysis of Human Judgments}
\label{sec:human-analysis}

To examine whether the proposed dimensions correspond to consistent and structured aspects of human judgment, we conducted a dense annotation study in which metaphor explanations were independently rated across all six dimensions and an additional holistic overall quality score. 
Analysis focuses on investigating the reliability, the structure, and the (dis)agreement between participants.

\subsection{Corpus and Annotation}
\label{ssec:corpus}

The corpus consists of $100$ metaphor explanations, each independently rated by $16$ annotators on all six dimensions and an overall quality score, using a five-point Likert scale, yielding a  dense corpus of $11,200$ individual human quality ratings ($16$ participants $\times$ $6+1$ dimensions $\times$ $100$ metaphor explanations).

\smallskip

\textbf{Metaphor Selection \& Explanation Generation.}\quad
The corpus was curated from linguistic databases, idiom collections, and natural language sources, covering diverse conceptual domains including physical, social, cultural, and abstract metaphors.
For each metaphor, one explanation was generated. The benchmark was deliberately engineered to span the full spectrum of explanation quality, comprising 15 reference explanations satisfying all six evaluation dimensions, 60 targeted failures violating exactly one dimension (10 per dimension), and 25 mixed failures violating multiple dimensions.

\smallskip

\textbf{Participants.}\quad
Each metaphor-explanation pair was independently evaluated according to detailed evaluation guidelines by 16 participants across the six proposed evaluation dimensions alongside an additional holistic overall quality score on a five-point Likert scale. 
Participants self-reported fluent-to-B2 English, spanning nine native languages, three reported linguistics/literature training, and ages range 18--73 (median = 21). All were recruited via personal contact.

\begin{table}[t]
\centering
\scriptsize
\definecolor{nordicbg}{RGB}{229,233,240}
\begin{tabular}{lccc}

\rowcolor{nordicbg}
\toprule
\textbf{Dimension} & \textbf{$\alpha$} & \textbf{ICC(2,1)} & \textbf{ICC(2,k)}\\
\midrule
\cellcolor{nordicbg}\textbf{Non-Circularity} & 0.370 & 0.377 & 0.906\\
\cellcolor{nordicbg}\textbf{Origin and Cultural\ Accuracy} & 0.423 & 0.433 & 0.924\\
\cellcolor{nordicbg}\textbf{Mapping} & 0.423 & 0.427 & 0.922\\
\cellcolor{nordicbg}\textbf{Emergence} & 0.579 & 0.583 & 0.957\\
\cellcolor{nordicbg}\textbf{Accessibility} & 0.456 & 0.460 & 0.932\\
\cellcolor{nordicbg}\textbf{Purpose} & 0.490 & 0.494 & 0.940\\
\midrule
\cellcolor{nordicbg}\textbf{Overall Quality} & 0.525 & 0.528 & 0.947\\
\bottomrule
\end{tabular}
\caption{Inter-rater reliability by dimension (100 metaphors, 16 raters).}
\label{tab:reliability}
\end{table}

\begin{table}[t]
\centering
\scriptsize
\definecolor{nordicbg}{RGB}{229,233,240}
\setlength{\tabcolsep}{4pt}
\begin{tabular}{lcccccc}
\rowcolor{nordicbg}
\toprule
 & \textbf{NC} & \textbf{OR} & \textbf{MP} & \textbf{EM} & \textbf{AC} & \textbf{PU}\\
\midrule
\cellcolor{nordicbg}\textbf{NC} & \cellcolor{blue!100}\textcolor{white}{\phantom{$-$}1.00} & \cellcolor{red!7}$-$0.07\phantom{$^{*}$} & \cellcolor{blue!12}0.12\phantom{$^{**}$} & \cellcolor{blue!13}0.13\phantom{$^{**}$} & \cellcolor{blue!0}0.00\phantom{$^{**}$} & \cellcolor{blue!17}0.17\phantom{$^{**}$}\\
\cellcolor{nordicbg}\textbf{OR} & \cellcolor{red!7}$-$0.07 & \cellcolor{blue!100}\textcolor{white}{\phantom{$-$}1.00\phantom{$^{*}$}} & \cellcolor{blue!0}0.00\phantom{$^{**}$} & \cellcolor{blue!1}0.01\phantom{$^{**}$} & \cellcolor{blue!21}0.21$^{*\phantom{*}}$ & \cellcolor{blue!2}0.02\phantom{$^{**}$}\\
\cellcolor{nordicbg}\textbf{MP} & \cellcolor{blue!12}\phantom{$-$}0.12 & \cellcolor{blue!0}\phantom{$-$}0.00\phantom{$^{*}$} & \cellcolor{blue!100}\textcolor{white}{1.00\phantom{$^{**}$}} & \cellcolor{blue!76}\textcolor{white}{0.76$^{**}$} & \cellcolor{blue!47}\textcolor{white}{0.47$^{**}$} & \cellcolor{blue!78}\textcolor{white}{0.78$^{**}$}\\
\cellcolor{nordicbg}\textbf{EM} & \cellcolor{blue!13}\phantom{$-$}0.13 & \cellcolor{blue!1}\phantom{$-$}0.01\phantom{$^{*}$} & \cellcolor{blue!76}\textcolor{white}{0.76$^{**}$} & \cellcolor{blue!100}\textcolor{white}{1.00\phantom{$^{**}$}} & \cellcolor{blue!49}\textcolor{white}{0.49$^{**}$} & \cellcolor{blue!89}\textcolor{white}{0.89$^{**}$}\\
\cellcolor{nordicbg}\textbf{AC} & \cellcolor{blue!0}\phantom{$-$}0.00 & \cellcolor{blue!21}\phantom{$-$}0.21$^{*}$ & \cellcolor{blue!47}\textcolor{white}{0.47$^{**}$} & \cellcolor{blue!49}\textcolor{white}{0.49$^{**}$} & \cellcolor{blue!100}\textcolor{white}{1.00\phantom{$^{**}$}} & \cellcolor{blue!54}\textcolor{white}{0.54$^{**}$}\\
\cellcolor{nordicbg}\textbf{PU} & \cellcolor{blue!17}\phantom{$-$}0.17 & \cellcolor{blue!2}\phantom{$-$}0.02\phantom{$^{*}$} & \cellcolor{blue!78}\textcolor{white}{0.78$^{**}$} & \cellcolor{blue!89}\textcolor{white}{0.89$^{**}$} & \cellcolor{blue!54}\textcolor{white}{0.54$^{**}$} & \cellcolor{blue!100}\textcolor{white}{1.00\phantom{$^{**}$}}\\
\bottomrule
\end{tabular}
\caption{Pairwise Spearman correlations ($^{*}p<.05$; $^{**}p<.0001$).}
\label{tab:correlations}
\end{table}

\begin{figure}[t]
\centering
  \includegraphics[width=0.78\columnwidth]{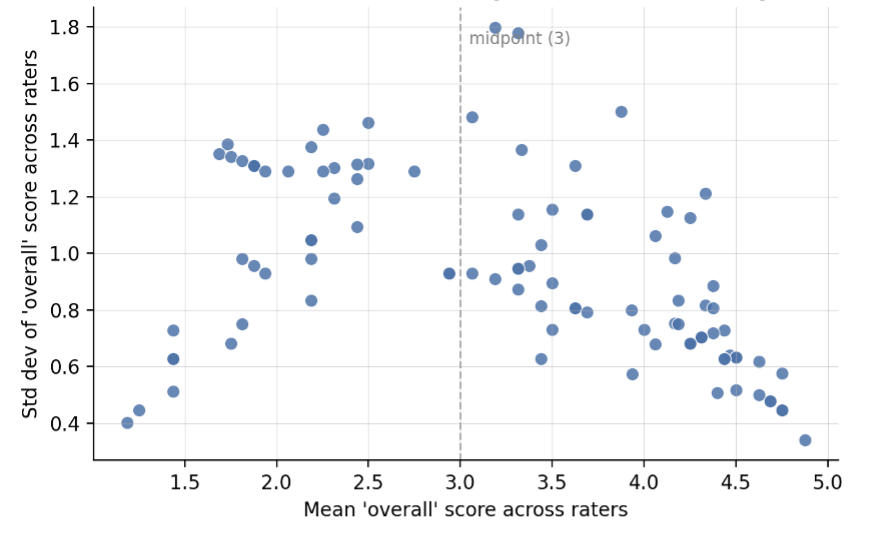}
  \caption{Mean score and rater deviation.}
  \label{fig:standard_deviation}
\end{figure}

\begin{table}[t]
\centering
\scriptsize
\definecolor{nordicbg}{RGB}{229,233,240}
\begin{tabular}{lrr} 
\rowcolor{nordicbg}
\toprule
\textbf{Predictor} & \textbf{Std.\ $\beta$} & \textbf{$p$} \\ 
\midrule
\cellcolor{nordicbg}\textbf{Non-Circularity} & 0.062 & <.001 \\ 
\cellcolor{nordicbg}\textbf{Origin and Cultural Accuracy} & $-$0.070 & <.001 \\ 
\cellcolor{nordicbg}\textbf{Mapping} & 0.061 & .004 \\ 
\cellcolor{nordicbg}\textbf{Emergence} & 0.499 & <.001 \\ 
\cellcolor{nordicbg}\textbf{Accessibility} & 0.014 & .420 \\ 
\cellcolor{nordicbg}\textbf{Purpose} & 0.290 & <.001 \\ 
\bottomrule
\end{tabular}
\caption{Standardized partial regression coefficients predicting Overall quality from the six criteria.}
\label{tab:partial-regression}
\end{table}

\subsection{Reliability, Structure, and Disagreement}

In the following we present analysis of inter-rater reliability, the correlational structure among dimensions, and the relationship between disagreement and explanation quality. Key results are:

\smallskip
\textbf{Agreement varies systematically across dimensions.} \quad
Krippendorff's $\alpha$ (interval) and ICC(2,1)/ICC(2,k) for each dimension (Table~\ref{tab:reliability}) shows that agreement varied substantially across dimensions, with Emergence showing the highest agreement ($\alpha=0.579$), while Non-Circularity shows the lowest ($\alpha=0.370$). 
No dimension reaches the conventional threshold for ``reliable'' agreement ($\alpha 0.8$), and most support only tentative conclusions. 
This suggests that annotators can apply these dimensions consistently enough to support the structural analyses, but not consistently enough to treat any single dimension's mean rating as a precise ground-truth signal in isolation. ICC(2,1) tracks $\alpha$ for every dimension, providing a cross-check that these estimates are not artifacts of the chosen reliability measure.


\smallskip
\textbf{The evaluation dimensions form two distinct groups.} \quad
Pairwise Spearman correlations between the six dimensions (Table~\ref{tab:correlations}) show that two dimensions (Non-Circularity and Origin \& Cultural Accuracy) are near-independent of every other dimension ($|\rho| \le 0.21$). In contrast, Mapping, Emergence, and Purpose form a strongly correlated cluster ($\rho = 0.76$--$0.89$, $p < 10^{-25}$), and Accessibility correlates moderately with this cluster ($\rho \approx 0.47$--$0.54$). This structure suggests that the six dimensions do not function as six equally distinct criteria of judgment. 
Mapping, Emergence, and Purpose capture closely related aspects of explanation quality, while Non-Circularity and Origin \& Cultural Accuracy behave as rather independent criteria.

\smallskip
\textbf{Emergence and Purpose dominate the prediction of overall quality.} \quad
As several dimensions are highly correlated, pairwise correlations may overestimate their independent contribution to the overall quality. Therefore we fit a linear model predicting Overall quality from all six standardized dimensions simultaneously (Table~\ref{tab:partial-regression}). Emergence ($\beta = 0.499$, $p < .001$) remained the strongest independent predictor, followed by Purpose ($\beta = 0.290$, $p < .001$). The remaining dimensions contributed only comparatively small effects, with Accessibility showing no significant independent association. These results suggest that Emergence and Purpose account for most of the independent predictive signal, while Mapping, Emergence, and Purpose remain partially redundant rather than representing fully independent criteria of judgment.

\smallskip
\textbf{Disagreement is structured.} \quad
Rating disagreement followed a clear U-shaped relationship with explanation quality (Figure \ref{fig:standard_deviation}). Annotators agreed most on explanations receiving consistently high or low ratings and disagreed most on borderline cases. For example, \textit{``Love is a Fine Wine''} received near-unanimous ratings, whereas \textit{``A Dead End''} produced the largest variation across annotators, suggesting that disagreement reflects ambiguity in explanation quality.

\section{Preliminary Automatic Evaluation}
\label{sec:automatic}

In the following we explore the question of whether a standard automatic evaluation pipeline can recover any of this structure.\footnote{This study serves as a feasibility check, we do not claim competitive automatic evaluation.}

\smallskip
\textbf{Pipeline.} \quad
We fine-tuned a BERT-base encoder to predict the six dimensions from the metaphor and its explanation. Predicted scores were combined into an overall quality estimate using a gradient-boosted judge model.
We evaluate using grouped $k$-fold cross-validation at the metaphor level (no metaphor appears in both training and test folds within a fold).

\begin{table}[t]
\centering
\scriptsize
\definecolor{nordicbg}{RGB}{229,233,240}
\begin{tabular}{lrrr}
\rowcolor{nordicbg}
\toprule
\textbf{CRITERION} & \textbf{MAE} & \textbf{$R^2$} & \textbf{Weight}\\
\midrule
\cellcolor{nordicbg}\textbf{Non-Circularity} & 0.208 & 0.194 & 0.020\\
\cellcolor{nordicbg}\textbf{Origin and Cultural Accuracy} & 0.263 & 0.129 & 0.044\\
\cellcolor{nordicbg}\textbf{Mapping} & 0.354 & 0.509 & 0.046\\
\cellcolor{nordicbg}\textbf{Emergence} & 0.343 & 0.755 & 0.746\\
\cellcolor{nordicbg}\textbf{Accessibility} & 0.260 & 0.336 & 0.045\\
\cellcolor{nordicbg}\textbf{Purpose} & 0.350 & 0.684 & 0.099\\
\midrule
\cellcolor{nordicbg}\textbf{OVERALL QUALITY} & 0.689 & 0.373 & --\\
\bottomrule
\end{tabular}
\caption{Criterion-level automatic evaluation. 
(GroupKFold split, 80 train / 20 test metaphors).}
\label{tab:automatic-eval}
\end{table}


\begin{table}[t]
\centering
\scriptsize
\definecolor{nordicbg}{RGB}{229,233,240}
\begin{tabular}{lc}
\rowcolor{nordicbg}
\toprule
\textbf{METRIC} & \textbf{VALUE}\\
\midrule
\cellcolor{nordicbg}\textbf{Pearson $r$ (\(\sigma\) vs.\ Error)} & 0.397 ($p=.083$)\\
\cellcolor{nordicbg}\textbf{Spearman $\rho$ (\(\sigma\) vs.\ Error)} & 0.203 ($p=.392$)\\
\cellcolor{nordicbg}\textbf{Within Noise Band (Error $\le\sigma$)} & 13/20 (65.0\%)\\
\cellcolor{nordicbg}\textbf{Mean / Median Gold-Adjusted Error} & 0.814 / 0.835\\
\bottomrule
\end{tabular}
\caption{Relationship between human disagreement and model error on 20 held-out test metaphors.}
\label{tab:gold-metric}
\end{table}


\subsection{Results}
\label{ssec:automatic-results}

Preliminary findings from the feasibility study are:

\smallskip
\textbf{Prediction performance reflects the multidimensional structure.} \quad
Table~\ref{tab:automatic-eval} reports per-dimension MAE, $R^2$, and judge-model weight. The pipeline predicts Emergence most accurately ($R^2=0.76$), followed by Purpose ($R^2=0.68$) and Mapping ($R^2=0.51$). Prediction performance is lower for Accessibility ($R^2=0.34$), Non-Circularity ($R^2=0.19$), and Origin \& Cultural Accuracy ($R^2=0.13$). Overall quality is predicted with MAE$=0.69$, $R^2=0.37$.

\smallskip
\textbf{Learned structure partially mirrors the human structure.} \quad
The judge model assigns most weight to Emergence and Purpose, the same two dimensions identified as the strongest independent predictors of Overall quality in the human analysis. Given that Emergence and Purpose are themselves strongly correlated ($\rho=0.89$) and are also the two best-predicted dimensions ($R^2=0.76$ and $0.68$, respectively), this suggests that the automatic evaluator captures the dominant structure underlying human judgments while assigning comparatively little weight to the remaining dimensions.

\smallskip
\textbf{Model error shows a non-significant association with human disagreement.} \quad
Table~\ref{tab:gold-metric} compares model error against human rating variability on the 20 held-out test metaphors. Model error correlates positively with human disagreement in the expected direction (Pearson $r=0.397$, $p=.083$; Spearman $\rho=0.203$, $p=.392$), but neither correlation reaches significance, and only 13 of 20 predictions (65\%) fall within one human standard deviation of the mean rating. The direction of the association is consistent with that hypothesis, however, it should be confirmed with a larger test set.

\section{Discussion and Outlook}\label{sec:discussion}

We have presented a cognitively motivated multidimensional framework for evaluating metaphor explanations. Through an empirical study, we also show how the dimensions, i.e., the six complementary cognitive constructs, exhibit distinct patterns of reliability, redundancy, and disagreement. Furthermore, an exploratory automatic evaluation pipeline partially recovered this structure. The diagnostic insight provided by this multidimensional evaluation, together with the finding that explanation quality is organized around a small number of dominant underlying dimensions, provides a basis for developing future automatic evaluation systems (e.g., utilizing LLM-as-a-judge  \cite{huang2025llmjudge}) that explicitly assess these dimensions before aggregating them into an overall evaluation, thereby producing judgments that are both more interpretable and more closely aligned with the multidimensional structure of human judgments. 
The observed structure of human judgments further suggests that understanding how different aspects of metaphor explanation contribute to perceived quality is a promising direction for developing cognitively informed evaluation models.

%
Because our benchmark deliberately includes criterion-specific failures, future work should validate the framework on naturally occurring explanations produced by humans and contemporary language models, and investigate whether the identified core dimensions generalize beyond metaphor explanation to other open-ended explanation tasks.

\pagebreak

\section*{Limitations}
Our findings should be interpreted in light of several limitations. The proposed framework is theory-informed but has only been evaluated on a deliberately constructed benchmark of metaphor explanations, rather than naturally occurring explanations produced by humans or contemporary language models. Moreover, while the annotation study provides evidence that the proposed dimensions capture meaningful aspects of human evaluation, it does not establish that they constitute an exhaustive or universal characterization of explanation quality. Finally, the automatic evaluation study is exploratory, using a relatively small dataset to investigate feasibility rather than to establish a state-of-the-art evaluation system.

\section*{Ethical Statement}
This work studies the evaluation of metaphor explanations and does not involve the deployment of language models in real-world decision-making. Participants provided informed consent and annotated metaphor explanations using anonymous ratings. The benchmark consists of linguistic examples and intentionally constructed explanation failures, none of which contain personal or sensitive information.


%
%

\appendix

\begin{table*}[t]
\centering
\small
\setlength{\tabcolsep}{5pt}
\definecolor{nordicbg}{RGB}{229,233,240}
\begin{tabular}{
    p{0.22\textwidth}
    p{0.54\textwidth}
    r
    r
}
\rowcolor{nordicbg}
\toprule
\textbf{Metaphor}
& \textbf{Evaluated explanation}
& \textbf{Overall}
& \textbf{SD} \\
\midrule
\cellcolor{nordicbg}\textcolor{gray!70!black}{\textit{An apology is the super glue of life.}}
& A sincere expression of regret repairs damaged relationships so effectively that it restores trust and reconnects people who had emotionally separated.
& \cellcolor{nordicbg}4.88
& \cellcolor{nordicbg}0.34 \\
\cellcolor{nordicbg}\textcolor{gray!70!black}{\textit{The elephant in the room.}}
& The metaphore describing a large trunked mammal that has somehow entered a domestic living space or office building.
& \cellcolor{nordicbg}1.19
& \cellcolor{nordicbg}0.40 \\
\cellcolor{nordicbg}\textcolor{gray!70!black}{\textit{She is an open book.}}
& This means comparing a female human being to a printed collection of bound pages that is not currently closed.
& \cellcolor{nordicbg}1.25
& \cellcolor{nordicbg}0.45 \\
\cellcolor{nordicbg}\textcolor{gray!70!black}{\textit{His bruised ego will take time to recover.}}
& Describes the painful blow to someone's self-esteem after a failure, requiring a period of emotional healing to regain confidence.
& \cellcolor{nordicbg}4.75
& \cellcolor{nordicbg}0.45 \\
\cellcolor{nordicbg}\textcolor{gray!70!black}{\textit{A Herculean task.}}
& According to the Twelve Labors of Hercules in Greek myth, it describes a task requiring superhuman strength or extraordinary effort.
& \cellcolor{nordicbg}4.75
& \cellcolor{nordicbg}0.45 \\
\bottomrule
\end{tabular}
\caption{
    Metaphors with the highest annotator agreement (lowest Overall-rating standard deviation).
}
\label{tab:high-agreement}
\end{table*}

\begin{table*}[t]
\centering
\small
\setlength{\tabcolsep}{5pt}
\definecolor{nordicbg}{RGB}{229,233,240}
\begin{tabular}{
    p{0.22\textwidth}
    p{0.54\textwidth}
    r
    r
}
\rowcolor{nordicbg}
\toprule
\textbf{Metaphor}
& \textbf{Evaluated explanation}
& \textbf{Overall}
& \textbf{SD} \\
\midrule
\cellcolor{nordicbg}\textcolor{gray!70!black}{\textit{A dead end.}}
& This refers to a stratified cryospheric accumulation where frozen atmospheric water creates a fibrous textile covering over the terrestrial surface.
& \cellcolor{nordicbg}3.19
& \cellcolor{nordicbg}1.80 \\
\cellcolor{nordicbg}\textcolor{gray!70!black}{\textit{Hitting a brick wall.}}
& This is a statement describing a person who experiences a sudden physical impact against a solid, rectangular masonry structure.
& \cellcolor{nordicbg}3.31
& \cellcolor{nordicbg}1.78 \\
\cellcolor{nordicbg}\textcolor{gray!70!black}{\textit{Life is a roller coaster.}}
& This phrase frames human existence as a sequence of events using the logic of an amusement park.
& \cellcolor{nordicbg}3.88
& \cellcolor{nordicbg}1.50 \\
\cellcolor{nordicbg}\textcolor{gray!70!black}{\textit{Journalism is literature in hurry.}}
& The rapid production of written narratives about current events, created under tight deadlines without the luxury of time.
& \cellcolor{nordicbg}3.06
& \cellcolor{nordicbg}1.48 \\
\cellcolor{nordicbg}\textcolor{gray!70!black}{\textit{Frozen with fear.}}
& Frozen with fear means feeling fear so intense that a person becomes frozen, unable to move or respond.
& \cellcolor{nordicbg}2.50
& \cellcolor{nordicbg}1.46 \\
\bottomrule
\end{tabular}
\caption{
    Metaphors with the lowest annotator agreement (highest Overall-rating standard deviation).
}
\label{tab:low-agreement}
\end{table*}

\section*{Supplementary Material}

The appendix provides supplementary analyses supporting the results
presented in the main paper. These include (\ref{app:human_analyses}) additional analyses of
annotator disagreement, complete regression diagnostics, (\ref{app:automatic-evaluation}) further
evaluation of the exploratory automatic-evaluation pipeline, (\ref{app:participants})
aggregate participant characteristics, and (\ref{app:reproducibility}) reproducibility details.

\section{Additional Human Analyses}
\label{app:human_analyses}

\subsection{Annotator Disagreement Across Quality Levels}
\label{sec:appendix-u-shape}

\begin{figure}[h]
    \centering
    \includegraphics[width=\linewidth]{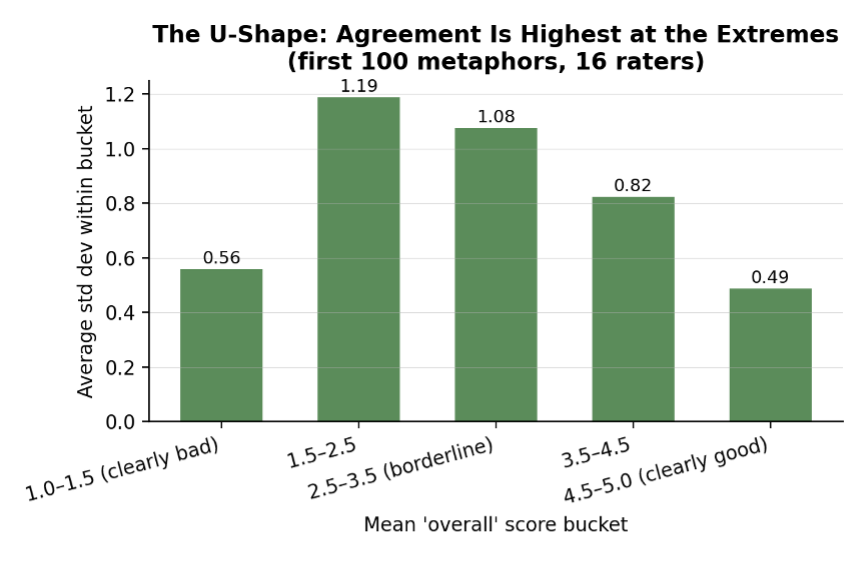}
    \caption{
        Average annotator disagreement across buckets of mean Overall quality. 
    }
    \label{fig:agreement-u-shape}
\end{figure}

Figure~\ref{fig:agreement-u-shape} summarizes annotator disagreement across buckets of mean Overall quality. Disagreement is operationalized as the standard deviation of the Overall ratings assigned to each metaphor. The resulting pattern is approximately U-shaped: annotators show greater agreement when an explanation is judged clearly poor or clearly successful, whereas disagreement is highest for explanations receiving intermediate scores.

\subsection{Disagreement Across Evaluation Dimensions}
\label{sec:appendix-criterion-disagreement}

\begin{figure}[h]
    \centering
    \includegraphics[width=\linewidth]{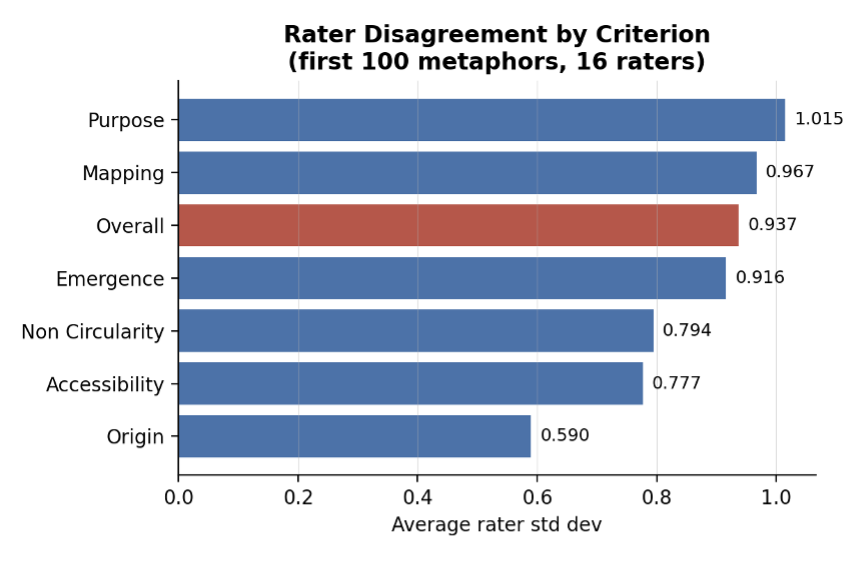}
    \caption{
        Average annotator disagreement by evaluation dimension. Bars represent the mean within-item standard deviation across metaphors.
    }
    \label{fig:criterion-disagreement}
\end{figure}

Figure~\ref{fig:criterion-disagreement} compares average annotator disagreement across the six evaluation dimensions and Overall quality. Purpose exhibits the greatest average variability, followed by Mapping. Origin exhibits the lowest variability, suggesting that annotators converged more strongly when evaluating the source or cultural basis of a metaphor than when evaluating its communicative purpose or conceptual mapping.

\subsection{Examples of High and Low Annotator Agreement}
\label{sec:appendix-agreement-examples}

Tables~\ref{tab:high-agreement} and~\ref{tab:low-agreement} present the five metaphors with the lowest and highest variability in Overall ratings, respectively. High agreement occurs at both extremes of the quality scale: annotators converge on explanations that are clearly successful as well as explanations that are clearly unsuccessful. The highest-disagreement cases generally receive intermediate mean scores, consistent with the aggregate pattern in Figure~\ref{fig:agreement-u-shape}.

\subsection{Full Regression Results}
\label{sec:appendix-regression}

\begin{table}[t]
\centering
\scriptsize
\setlength{\tabcolsep}{4pt}
\definecolor{nordicbg}{RGB}{229,233,240}
\begin{tabular}{lrrrrr}
\rowcolor{nordicbg}
\toprule
\textbf{Pred.}
& \textbf{$\beta$}
& \textbf{SE}
& \textbf{$t$}
& \textbf{$p$}
& \textbf{90\% CI} \\
\midrule
\cellcolor{nordicbg}\textbf{NC}
& 0.062
& 0.016
& 3.91
& $<.001$
& [0.031, 0.094] \\
\cellcolor{nordicbg}\textbf{OR}
& $-0.070$
& 0.016
& $-4.42$
& $<.001$
& [$-0.102$, $-0.039$] \\
\cellcolor{nordicbg}\textbf{MP}
& 0.061
& 0.021
& 2.90
& .004
& [0.020, 0.103] \\
\cellcolor{nordicbg}\textbf{EM}
& 0.499
& 0.023
& 21.98
& $<.001$
& [0.454, 0.543] \\
\cellcolor{nordicbg}\textbf{AC}
& 0.014
& 0.017
& 0.81
& .420
& [$-0.020$, 0.047] \\
\cellcolor{nordicbg}\textbf{PU}
& 0.290
& 0.023
& 12.51
& $<.001$
& [0.245, 0.335] \\
\bottomrule
\end{tabular}
\caption{
    Full standardized regression output predicting Overall quality
    from all six evaluation dimensions simultaneously. Coefficients
    represent associations with Overall quality while holding the
    remaining dimensions constant.
}
\label{tab:full-regression}
\end{table}

Table~\ref{tab:full-regression} reports complete simultaneous regression predicting Overall quality from the six standardized evaluation dimensions. Emergence provides the strongest independent predictive signal, followed by Purpose. The remaining coefficients are comparatively small, and Accessibility does not exhibit a significant independent association after controlling for the other dimensions.
The small negative coefficient for Origin should be interpreted cautiously. Its sign may reflect shared variance or a suppression effect among correlated predictors rather than a substantively negative relationship between origin accuracy and explanation quality.

\section{Additional Analyses of the Automatic-Evaluation}
\label{app:automatic-evaluation}

\subsection{Criterion-Level Generalization}
\label{sec:appendix-criterion-generalization}

Table~\ref{tab:criterion-generalization} compares mean absolute error (MAE) on the training and held-out test sets for each evaluation dimension. Most dimensions exhibit higher error on the held-out metaphors, indicating a degree of overfitting consistent with the modest size of the available dataset. The increase is largest for Purpose and Mapping, suggesting that these dimensions are comparatively more difficult to generalize. By contrast, Non-Circularity shows slightly lower error on the test set, although this difference is small and should not be overinterpreted given the limited number of held-out examples.

\begin{table}[h]
\centering
\small
\definecolor{nordicbg}{RGB}{229,233,240}
\begin{tabular}{lrrr}
\rowcolor{nordicbg}
\toprule
\textbf{Dimension}
& \textbf{Train MAE}
& \textbf{Test MAE}
& \textbf{$\Delta$} \\
\midrule
\cellcolor{nordicbg}\textbf{NC}
& 0.261
& 0.208
& $-0.053$ \\
\cellcolor{nordicbg}\textbf{OR}
& 0.243
& 0.263
& 0.020 \\
\cellcolor{nordicbg}\textbf{MP}
& 0.252
& 0.354
& 0.102 \\
\cellcolor{nordicbg}\textbf{EM}
& 0.251
& 0.343
& 0.092 \\
\cellcolor{nordicbg}\textbf{AC}
& 0.210
& 0.260
& 0.049 \\
\cellcolor{nordicbg}\textbf{PU}
& 0.239
& 0.350
& 0.110 \\
\bottomrule
\end{tabular}
\caption{
    Training and held-out-test MAE for the six criterion-level
    predictors. Positive values of $\Delta$ indicate higher error
    on the held-out metaphors.
}
\label{tab:criterion-generalization}
\end{table}

\subsection{Association Between Human Variability and Model Error}
\label{sec:appendix-error-correlation}

The relationship between human disagreement and automatic-evaluation error was positive but uncertain. Pearson correlation produced $r=.397$ ($p=.083$), while Spearman correlation produced $\rho=.203$ ($p=.392$). Thus, although the observed direction is consistent with the possibility that examples that are difficult for humans are also difficult for the automatic evaluator, the evidence does not establish a reliable association.

\subsection{High- and Low-Disagreement Groups}
\label{sec:appendix-disagreement-groups}

As an additional exploratory analysis, the held-out metaphors were divided using a median split of their Overall-rating standard deviation ($\sigma=0.935$).

\begin{table}[h]
\centering
\small
\definecolor{nordicbg}{RGB}{229,233,240}
\begin{tabular}{lrrrr}
\rowcolor{nordicbg}
\toprule
\textbf{Group}
& \textbf{$n$}
& \makecell{\textbf{Mean}\\ \textbf{SD}}
& \makecell{\textbf{Mean}\\ \textbf{error}}
& \makecell{\textbf{Median}\\ \textbf{error}} \\
\midrule
\cellcolor{nordicbg}\textbf{High disagreement}
& 10
& 1.151
& 0.790
& 0.700 \\
\cellcolor{nordicbg}\textbf{Low disagreement}
& 10
& 0.636
& 0.584
& 0.525 \\
\bottomrule
\end{tabular}
\caption{
    Automatic-evaluation error for held-out metaphors with high and
    low human-rating disagreement.
}
\label{tab:high-low-disagreement}
\end{table}

Although error was descriptively greater among the high-disagreement examples, the difference was not statistically significant under either a Mann--Whitney test ($U=58.00$, $p=.571$) or Welch's $t$-test ($t=1.07$, $p=.303$). The comparison is exploratory given the small number of held-out metaphors.

\section{Participant Information}
\label{app:participants}

Participants (Table \ref{tab:participant-demographics}) represented a range of linguistic, cultural, and educational backgrounds. The study recorded native or first language, self-reported English proficiency, cultural or geographic background, familiarity with linguistics, literature, or metaphor, age or educational band, and recruitment method.

\begin{table}[h]
\centering
\small
\definecolor{nordicbg}{RGB}{229,233,240}
\begin{tabular}{p{0.50\linewidth}p{0.38\linewidth}}
\rowcolor{nordicbg}
\toprule
\textbf{Characteristic}
& \textbf{Sample description} \\
\midrule
\cellcolor{nordicbg}\textbf{Included annotators}
& 16 \\
\cellcolor{nordicbg}\textbf{English proficiency}
& B2 or fluent \\
\cellcolor{nordicbg}\textbf{Number of native languages}
& 9 \\
\cellcolor{nordicbg}\textbf{Age range}
& 18--73 \\
\cellcolor{nordicbg}\textbf{Median age}
& 21 \\
\cellcolor{nordicbg}\textbf{Linguistics or literature training}
& 3 participants \\
\cellcolor{nordicbg}\textbf{Recruitment method}
& Personal contact \\
\bottomrule
\end{tabular}
\caption{Aggregate characteristics of the annotator sample.}
\label{tab:participant-demographics}
\end{table}

\section{Reproducibility Details}
\label{app:reproducibility}

\subsection{Annotation Protocol}

\begin{figure*}[t]
    \centering
    \includegraphics[width=0.8\linewidth]{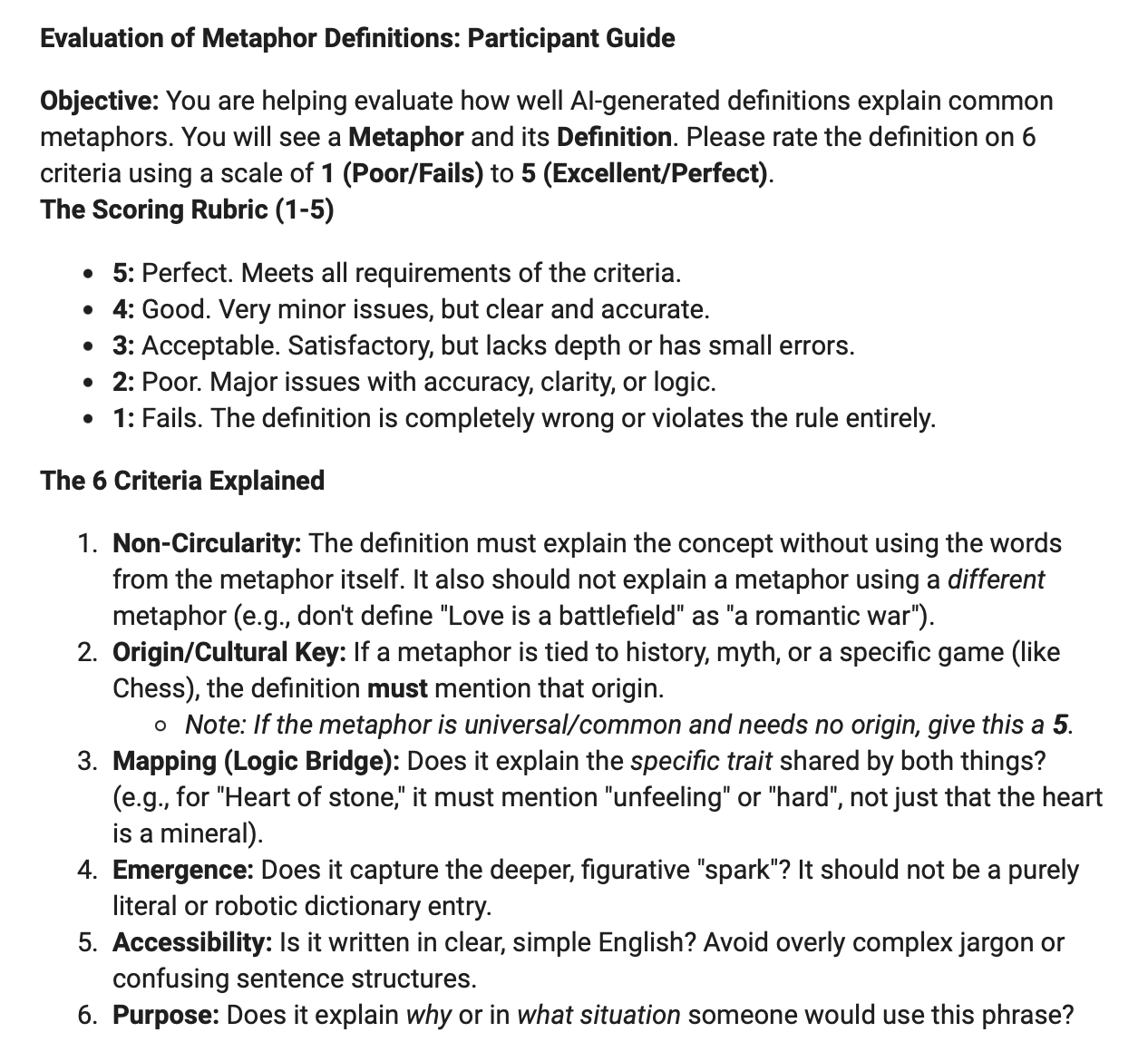}
    \caption{Participant instructions reproduced as displayed in the Google
    Forms survey.}
    \label{fig:participant-instructions}
\end{figure*}

Participants completed the annotation task through a Google Forms survey. The survey header presented the task objective, the 1--5 scoring rubric, and the six evaluation criteria (Figure~\ref{fig:participant-instructions}). 
The survey was completed independently and asynchronously. Participants did not complete the task in a single supervised session and did not communicate with one another during annotation. Responses were exported from Google Forms as a CSV file for analysis.

\subsection{Explanation Generation and Benchmark Construction}

The benchmark contains 100 metaphor explanations drawn from a range of physical,
social, cultural, and abstract conceptual domains. Source metaphors were curated
from publicly available metaphor collections and supplemented with idiomatic
expressions compiled by the authors from general usage.
Candidate explanations were generated through the web interfaces of
three commercial large language models: GPT-4o, Claude 3.5 Sonnet, and Gemini
1.5 Pro. A standardized prompt template was used across systems. Explanations
were normalized to approximately 25 words and to a broadly uniform,
professional tone.

The generation prompt was based on the following template:

\begin{quote}
\textbf{Role:} You are a computational linguistics expert.

\textbf{Task:} Provide a definition for the metaphor: ``Frozen with fear.''

\textbf{Constraint 1:} The definition must be approximately 25 words.

\textbf{Constraint 2, targeted or mixed failures:} The definition must
intentionally fail the specified subset of criteria while passing all others.

\textbf{Constraint 2, reference explanations:} The definition must pass all six
criteria.

\end{quote}

The following criteria definitions were provided:

\begin{enumerate}
\item \textbf{Non-Circularity:} The definition must not restate the metaphor using its own words, or explain it via another metaphor.
\item \textbf{Origin and Cultural Accuracy:} The definition must identify the specific cultural, historical, or domain-specific origin of the metaphor (e.g., Chess, Myth, Navy), where applicable. This criterion applies only when the metaphor's source or target domain is traceable to a specific cultural tradition, historical event, game, field, or institution; a fail means the definition describes the meaning but ignores a traceable source.
\item \textbf{Emergence:} The definition must capture the figurative meaning that emerges from the metaphor, not merely describe the literal, physical objects involved.
\item \textbf{Accessibility:} The definition must use clear, natural language, avoiding dense academic jargon.
\item \textbf{Purpose:} The definition must explain the emotional or situational intent behind using the phrase (why or when someone would use it).
\item \textbf{Mapping:} The definition must identify the specific trait shared by the source and target domains (e.g., “vastness,” “heat,” “sharpness”), rather than defining the target without the linking trait.
\end{enumerate}

Generated explanations were manually checked against the intended quality
profile. Outputs that were hallucinated, off target, or inconsistent with the
requested failure pattern were filtered and regenerated until the intended
benchmark distribution was obtained.

The final benchmark contains:

\begin{itemize}
    \item \textbf{15 reference explanations} designed to satisfy all six criteria;
    \item \textbf{60 targeted failures}, with ten explanations designed to violate
    each individual criterion; and
    \item \textbf{25 mixed failures} designed to violate multiple criteria
    simultaneously.
\end{itemize}

\subsection{Automatic Evaluation Pipeline}

The automatic evaluator consists of two stages. The first stage predicts the six
criterion scores and the overall human judgment from the metaphor--explanation
pair. The second stage maps the six predicted criterion scores to a final overall
quality estimate.

\smallskip

\textbf{Stage 1: Criterion Prediction}

We fine-tuned \texttt{bert-base-uncased} as a multi-output regression model. The
metaphor and candidate explanation were concatenated into a single input
sequence, allowing the encoder to model the relation between the expression and
its proposed explanation. A linear regression head mapped the encoder
representation to seven continuous targets: the six criterion scores and the
overall quality score.

BERT was selected as a computationally lightweight alternative to an LLM-based
judge and to reduce the risk of self-preference effects that may arise when a
generative model evaluates outputs produced by similar systems.

\smallskip

\textbf{Stage 2: Overall-Score Prediction}

The six criterion predictions produced by Stage 1 were used as input features to
a Gradient Boosting regressor implemented in scikit-learn. The model predicts the
final overall quality score.
Gradient Boosting was used through five-fold \texttt{GroupKFold}
cross-validation on the training portion, with grouping performed by metaphor to
avoid leakage between explanations associated with the same metaphor. The model
used 200 boosting estimators. Hyperparameters are reported in Table \ref{tab:gb-hyperparameters}.

\begin{table}[t]
\centering
\small
\definecolor{nordicbg}{RGB}{229,233,240}
\begin{tabular}{ll}
\rowcolor{nordicbg}
\toprule
\textbf{Hyperparameter} & \textbf{Value} \\
\midrule
\cellcolor{nordicbg}\textbf{Estimator} & \texttt{GradientBoostingRegressor} \\
\cellcolor{nordicbg}\textbf{Number of estimators} & 200 \\
\cellcolor{nordicbg}\textbf{Learning rate} & scikit-learn default \\
\cellcolor{nordicbg}\textbf{Maximum tree depth} & scikit-learn default \\
\cellcolor{nordicbg}\textbf{Subsample} & scikit-learn default \\
\cellcolor{nordicbg}\textbf{Loss} & scikit-learn default \\
\cellcolor{nordicbg}\textbf{Random state} & 42\\
\bottomrule
\end{tabular}
\caption{Stage-2 Gradient Boosting configuration.}
\label{tab:gb-hyperparameters}
\end{table}

\subsection{Training Configuration}

\begin{table}[t]
\centering
\small
\definecolor{nordicbg}{RGB}{229,233,240}
\begin{tabular}{ll}
\rowcolor{nordicbg}
\toprule
\textbf{Hyperparameter} & \textbf{Value} \\
\midrule
\cellcolor{nordicbg}\textbf{Base checkpoint} & \texttt{bert-base-uncased} \\
\cellcolor{nordicbg}\textbf{Task head} & \texttt{BertForSequenceClassification} \\
\cellcolor{nordicbg}\textbf{Number of labels} & 7 \\
\cellcolor{nordicbg}\textbf{Problem type} & Regression \\
\cellcolor{nordicbg}\textbf{Loss function} & Mean squared error \\
\cellcolor{nordicbg}\textbf{Optimizer} & AdamW \\
\cellcolor{nordicbg}\textbf{Learning rate} & $2\times 10^{-5}$ \\
\cellcolor{nordicbg}\textbf{Training batch size} & 2 \\
\cellcolor{nordicbg}\textbf{Evaluation batch size} & Hugging Face default \\
\cellcolor{nordicbg}\textbf{Weight decay} & 0.01 \\
\cellcolor{nordicbg}\textbf{Epochs} & 50, fixed \\
\cellcolor{nordicbg}\textbf{Early stopping} & None \\
\cellcolor{nordicbg}\textbf{Dropout} & 0.1, BERT default \\
\cellcolor{nordicbg}\textbf{Max. sequence length} & 128 tokens \\
\cellcolor{nordicbg}\textbf{Data-split seed} & 42 \\
\cellcolor{nordicbg}\textbf{Training/init. seed} & 42 \\
\cellcolor{nordicbg}\textbf{Hardware} & Apple M2 \\
\cellcolor{nordicbg}\textbf{Backend} & PyTorch MPS, with CPU fallback \\
\cellcolor{nordicbg}\textbf{Approx. training time} & 3 hours \\
\bottomrule
\end{tabular}
\caption{Hyperparameters used to fine-tune the Stage-1 BERT regressor.}
\label{tab:bert-hyperparameters}
\end{table}

Table~\ref{tab:bert-hyperparameters} reports the Stage-1 training settings.
Hyperparameters were set manually using standard defaults for BERT fine-tuning
on a small regression dataset.
The model was trained on 1,600 annotation instances derived from 16 annotators and 100 metaphor explanations. Train--test splitting was performed at the metaphor level to prevent leakage. The held-out evaluation set contained 20 metaphors.

\pagebreak

\subsection{Reproducibility Checklist}

\begin{itemize}
    \item \textbf{Dataset size:} 100 metaphor explanations.
    \item \textbf{Annotators:} 16.
    \item \textbf{Annotation targets:} six criterion scores plus overall quality.
    \item \textbf{Split strategy:} metaphor-level train--test split.
    \item \textbf{Held-out set:} 20 metaphors.
    \item \textbf{Stage-1 model:} \texttt{bert-base-uncased} with a seven-output
    regression head.
    \item \textbf{Stage-2 model:} scikit-learn Gradient Boosting regressor.
    \item \textbf{Primary software:} PyTorch, Hugging Face Transformers, and
    scikit-learn.
    \item \textbf{Hardware:} Apple M2 using the PyTorch MPS backend, with CPU
    fallback where required.
    \item \textbf{Reported runs:} one train--test split and one training run, not
    an average across random seeds.
    \item \textbf{Code and data availability:} The benchmark, human annotations, and source code is available upon request by contacting: \href{mailto:info@codesign-lab.org}{info@codesign-lab.org}
\end{itemize}

\end{document}